\documentclass[letterpaper]{article} 
\usepackage{aaai24}  
\usepackage{times}  
\usepackage{helvet}  
\usepackage{courier}  
\usepackage[hyphens]{url}  
\usepackage{graphicx} 
\usepackage{natbib}  
\usepackage{caption} 
\usepackage{algorithm}
\usepackage{algorithmic}

\usepackage{graphicx}
\usepackage{amsmath}
\usepackage{amsfonts,amssymb}

\usepackage{booktabs}
\usepackage{multirow}

\usepackage{newfloat}
\usepackage{listings}
\DeclareCaptionStyle{ruled}{labelfont=normalfont,labelsep=colon,strut=off} 
\floatstyle{ruled}
\newfloat{listing}{tb}{lst}{}
\floatname{listing}{Listing}
\title{Pseudo-Label Calibration Semi-supervised Multi-Modal Entity Alignment}
\author{
    Luyao Wang, Pengnian Qi, Xigang Bao, Chunlai Zhou, Biao Qin\thanks{Corresponding author.}
}
\affiliations{
    School of Information, Renmin University of China\\

    \{wangluyao123, pengnianqi, baoxigang, czhou, qinbiao\} @ruc.edu.cn
}

\usepackage{bibentry}

\begin{document}

\maketitle

\begin{abstract}
Multi-modal entity alignment (MMEA) aims to identify equivalent entities between two multi-modal knowledge graphs for integration. Unfortunately, prior arts have attempted to improve the interaction and fusion of multi-modal information, which have overlooked the influence of modal-specific noise and the usage of labeled and unlabeled data in semi-supervised settings. In this work, we introduce a Pseudo-label Calibration Multi-modal Entity Alignment (PCMEA) in a semi-supervised way. Specifically, in order to generate holistic entity representations, we first devise various embedding modules and attention mechanisms to extract visual, structural, relational, and attribute features. Different from the prior direct fusion methods, we next propose to exploit mutual information maximization to filter the modal-specific noise and to augment modal-invariant commonality. Then, we combine pseudo-label calibration with momentum-based contrastive learning to make full use of the labeled and unlabeled data, which improves the quality of pseudo-label and pulls aligned entities closer. Finally, extensive experiments on two MMEA datasets demonstrate the effectiveness of our PCMEA, which yields state-of-the-art performance. 
\end{abstract}

\section{Introduction}

Multi-modal knowledge graphs (MMKGs) have drawn massive attention in various scenarios and motivated numerous downstream applications \cite{sun2020multi,ding2022mukea,shao2023prompting}. In MMKGs, knowledge is often summarized in various forms, such as relation triples, attribute triples, and images. Generally, MMKGs are constructed for specific purposes, leading to separate MMKGs with different descriptions for identical concepts. To improve the completeness of MMKGs, multi-modal entity alignment (MMEA) is an emerging tasks that link entities referring to the same real-world concept.

\begin{figure}[h!]
	\centering
	\includegraphics[width=\linewidth]{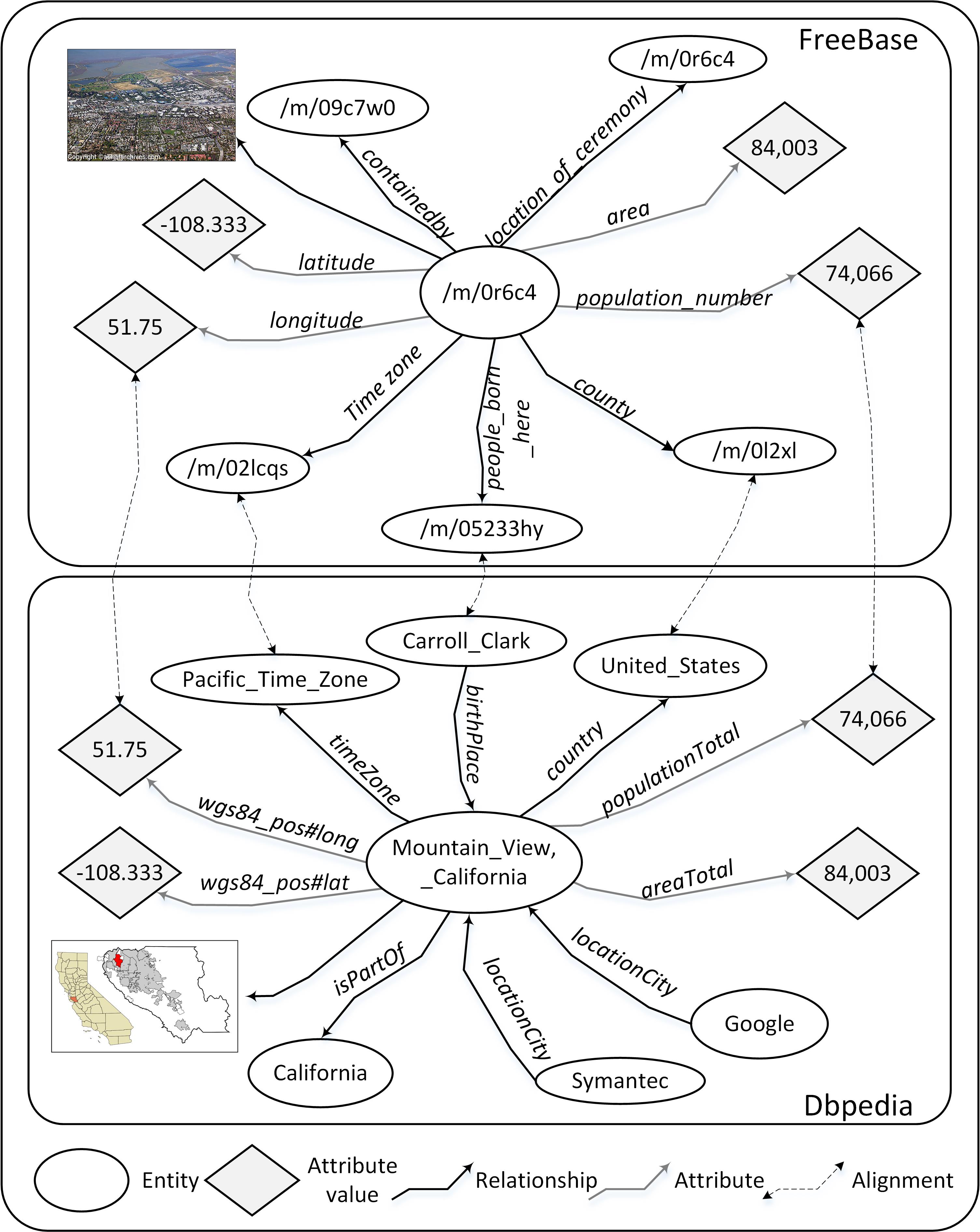}
	\caption{An example of multi-modal entity alignment. The oval shape represents entities, and the diamond shape represents attribute values. The dotted line indicates the relation or attribute of the alignment of two aligned entities. }
	\label{Intro}
\end{figure}

Figure \ref{Intro} illustrates a toy example in MMEA. Commonly, aligned entities share similarities in attributes, relations, topology, or visual information. Thus, recent works \cite{chen2020mmea, lin2022multi} design interaction and fusion methods to integrate multi-modal embeddings. Nevertheless, 1) direct interaction and fusion introduce richer information and modal-specific noise. For instance, the entity \textit{/m/0r6c4} in FreeBase and the entity \textit{Mountain\_View,\_California} in Dbpedia have similar relation (e.g. “Time zone” vs. “timeZone”) and attribute (e.g. “population\_number” vs. “populationTotal”), all of which favor the alignment of the two entities. But these entities may be incorrectly aligned due to significant visual differences. 2) The exploration of optimal embedding methods for each modality is often neglected. For example, “longitude” and “wgs84\_pos\#long” represent the full name and abbreviation, and the bag-of-words method might maximize similarity. Conversely, embedding by the pre-trained language model might be better for the case where “birthplace” and “people\_born\_here” are different phrases with the same meaning. In addition, most existing methods exploit embedding-based approaches relying heavily on human labeling. However, 3) the design of the training strategy with limited labels has been overlooked in MMEA. Many methods only use existing labels for supervised learning, neither fully utilizing unlabeled data nor preventing model bias.

\textbf{Contributions:} To address the abovementioned  problems, we introduce PCMEA, a Pseudo-label Calibration based semi-supervised MMEA framework. It has three main components: PCMEA first utilizes diverse encoders and attention mechanisms to obtain modality-specific representations for each entity. To exploit complementarities across modalities and filter out model-specific noise, PCMEA then employs mutual information-enhanced cross-modality alignment methods, which can enrich intra-modal interaction and avoid the influence of noise. To leverage labeled and unlabeled data, PCMEA finally develops momentum-based contrastive learning with pseudo-label calibration, which can reduce error propagation and help to align entities. Experimental results show that our approach achieves state-of-the-art performance on two MMEA benchmark datasets.

\section{Related Work}

\textbf{Entity alignment (EA) and multi-modal entity alignment.} With developments in knowledge graph representation learning, embedding-based entity alignment has emerged. Those embedding-based methods commonly have two steps: 1)KG embedding module encodes the entities into vectors according to the semantic or structural information; 2)entity alignment \cite{sun2020benchmarking} module captures the correspondence of embedding vectors with seed alignment. 

Recently, lots of multi-modal knowledge graphs have been constructed. Information from different independent modalities can complement each other. Nevertheless, their corresponding representations, reflected in separate spaces, can not directly merge in a shared space. A fusion module in MMEA \cite{chen2020mmea} is proposed to migrate knowledge embeddings from multiple modalities for robustness. A novel method in MSNEA \cite{chen2022multi} uses visual features to guide other modalities learning, promoting inter-modal enhanced entity representation. In addition to the modal embedding perspective, an intra-modal contrastive loss in MCLEA \cite{lin2022multi} is utilized to distinguish the embeddings of equivalent entities from other entities for each modality.

Different from previous methods, our proposed PCMEA not only strengthens the complementarity between different modal representations using mutual information and attention mechanism but also devises a more effective contrastive learning strategy to pull aligned entities closer.

\textbf{Semi-supervised and unsupervised entity alignment.}
Recently, some EA methods based on representation learning have been prevalent due to their accuracy. However, their success often relies on annotated data, which means higher labor costs. For situations with few or no aligned seeds, several semi-supervised or unsupervised EA approaches have been proposed. Methods like CEAFF \cite{zeng2021reinforcement} and RLEA \cite{guo2022deep} transform the alignment process into sequence decision task. RAC \cite{zeng2021reinforced} conducts reinforced active learning by selecting entities to manually label with minimal labeling efforts and exploit vast unlabeled data. Meanwhile, self-supervised EA methods like EVA \cite{liu2021visual} and ICLEA \cite{zeng2022interactive} create pre-aligned entity pairs by leveraging visual similarity or contrastive learning. However, those semi-supervised or self-supervised methods usually suffer from error accumulation, leading to performance bottlenecks. The main reason is no guarantee of the accuracy of the pre-decision or pre-alignments. Therefore, in PCMEA, we calibrate the pseudo-labels before incorporating them into the supervised contrastive learning framework, decreasing error accumulation.

\begin{figure*}[t]
	\centering
	\includegraphics[width=\linewidth]{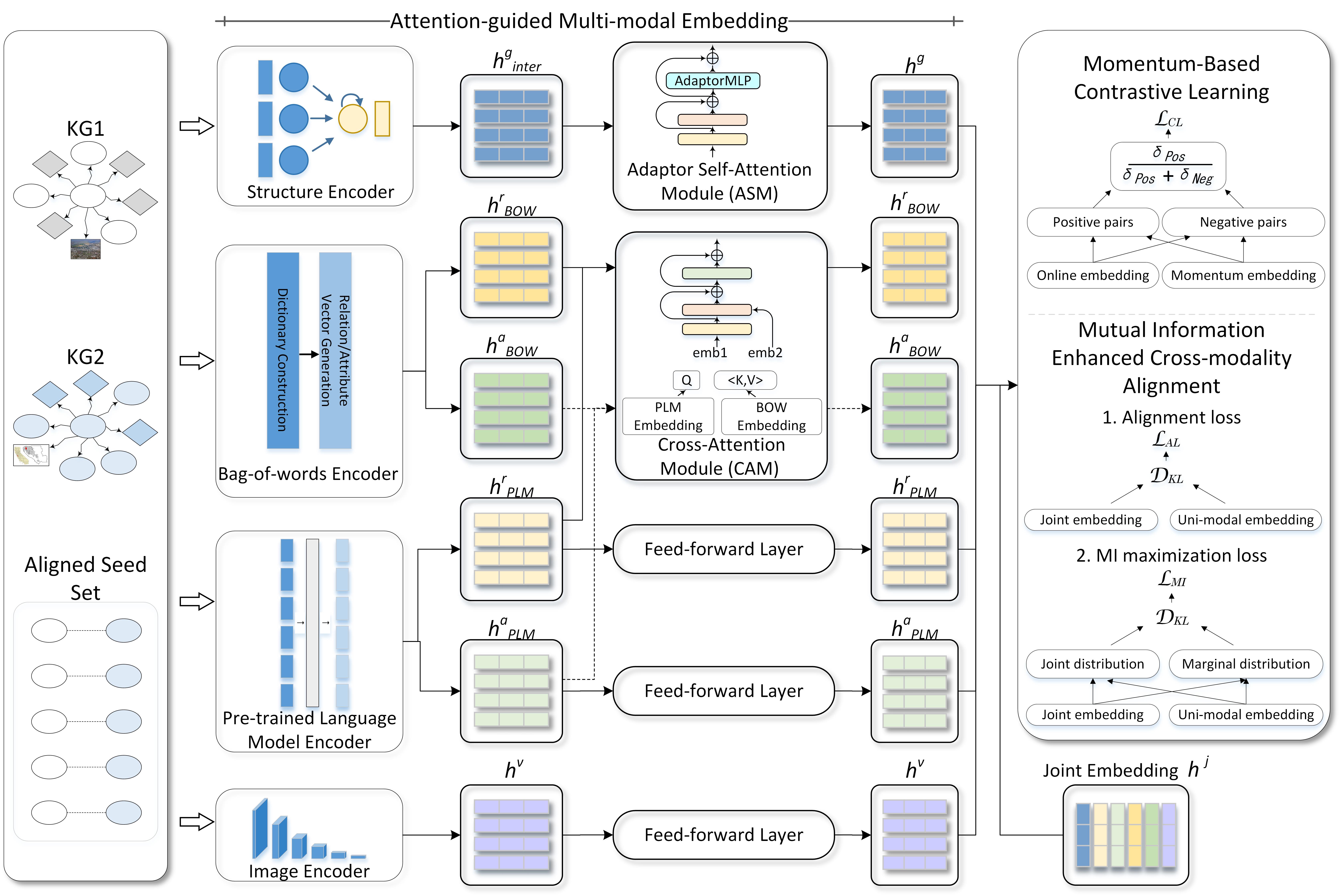}
	\caption{The overall architecture of PCMEA, which combines heterogeneous multi-modal attention-guided embedding and learns through MI maximization enhanced alignment loss (consists of alignment loss $\mathcal {L}_{AL}$ and MI maximization loss $\mathcal {L}_{MI}$), and momentum-based contrastive loss $\mathcal {L}_{CL}$.}
	\label{frame}
\end{figure*}

\section{Methodology}

\subsection{Problem Definition and Notations}
\textit{Definition 1:Multi-modal Knowledge Graph.} 
A multi-modal knowledge graph is formalized as $G = (E, R, I, A, V, T_R, T_A)$. Here, $E, R, I, A $, and $ V$ denote the sets of entities, relations, images, attributes, and values, respectively. $T_R = \{(h,r,t) | h,t\in E, r \in R\}$ refers to the set of relation triples. $T_A = \{(e, a, v)| e\in E, a \in A, v \in V\}$ denotes the set of attribute triples.

\textit{Definition 2:Multi-modal Entity alignment.} 
Given two multi-modal knowledge graphs $G$ and $G'$, $G$=$ (E, R, I, A, V, T_R, T_A)$ and $G'$=$ (E', R', I', A', V',$ $T_{R}', T_{A}')$, the set of alignment seeds across two multi-modal knowledge graphs is defined as $H = \{(e, e')| e\in E, e'\in E', e\equiv e'\}$,
where $\equiv$ represents the equivalence of two entities. The task of multi-modal entity alignment targets to match the counterpart entities $e$ and $e'$, which describe the same concepts in the real world from distinct multi-modal knowledge graphs.

\subsection{Framework Overview}
In this paper, we introduce a semi-supervised multi-modal entity alignment framework called PCMEA to solve the challenges above. Our proposed PCMEA comprises three components: \textit{Attention-guided Multi-modal Embedding}  to extract visual, relation, attribute, and structure features with diverse encoders and attention mechanisms; \textit{Mutual Information Enhanced Cross-modality Alignment} to encourage cross-modality knowledge transfer and to filter modality-invariant noise; \textit{Contrastive Learning with Pseudo-Label Calibration} method to help align entities with a few label supervision. 

\subsection{Attention-guided Multi-modal Embedding}

In multi-modal knowledge graphs, there are various modalities of knowledge to depict an entity, i.e., neighborhood structure, relations, attributes, and images. Each modality is processed using different encoders depending on the nature of the signal. Furthermore, uni-modal embeddings are fused with weighted concatenation to form the joint embedding. 

\subsubsection{Self-attention Augmented Structure Embedding.}
We utilize Graph Attention Networks (GAT) \cite{velivckovic2018graph} to model the neighborhood structure information of entities as real-valued vectors, since GAT aggregates neighborhood information with the attention mechanism and focuses on the most relevant neighbors. In practice, we apply a two-layer GAT model to embed the neighborhood information. We use the output of the last layer as the intermediate representation of neighbor structure embedding $h_{inter}^{g}$.

Because diverse modalities have different statistical properties, which are distributed cross feature spaces, we apply an adaptor self-attention module (ASM) to narrow the semantic gap. The main components of ASM are a multi-head self-attention (MHSA) layer and a plug-and-play bottleneck layer called AdaptorMLP \cite{chen2022adaptformer}. The processes of $h^g_{inter}$ and $h^g$ are formulated as:
\begin{align}
&h_{inter}^{g} = W_g \cdot \mathit{GAT}_2(g_i)+b_g \\
&h^{g} = \mathit{ASM}(h_{inter}^{g})
\end{align}
where $W_g$ and $b_g$ are learnable parameters.

\subsubsection{Semantic Information Guided Relation and Attribute Embedding.}
In KGs, the relation triples (attribute triples) of the corresponding entities have similarities in characters or semantics. Thus, we adopt two approaches to embed the relations (attributes) triples. On one hand, we represent the relations (attributions) of entities $e_i$ as bag-of-words features and feed them into a feed-forward layer to obtain the relation embedding $h_{BOW}^{r}$ (attribute embedding $h_{BOW}^{a}$). On the other hand, we expand the relation (attribute) triples $t_m$ into word sequences $s_m$ and input $s_m$ into a pre-trained language model, which can understand the meaning of sentences. In our work, we apply T5 \cite{raffel2020exploring} and Roberta \cite{liu2019roberta} to encode relation and attribute triples, respectively. After a feed-forward layer, the semantic representation of relation (attribute) triples can be obtained.
\begin{align}
&h^{m}_{BOW} = W_m \cdot \mathit{BOW}(t_m)+b_m, m\in\{r,a\} \\
&h^{m}_{PLM} = \mathit{PLM}(s_m), m\in\{r,a\} \\
&h^{m}_{PLM} = W_m' \cdot h^m_{PLM} +b_m', m \in \{r,a\}
\end{align}
where $W_m$, $W_m'$, $b_m$, and $b_m'$ are learnable parameters.

The character and semantic information of relation triples and attribute triples can be mutually complemented. Therefore, we adopt a cross-attention module and use semantic information to guide further learning of character information. The cross-attention module (CAM) is very similar to multi-head self-attention, with the difference that semantic embeddings $h_{\mathit{PLM}}^{m}$ are used as query inputs and character embeddings $h^{m}_{\mathit{BOW}}$ are used as the input of keys and values. The structure of CAM is shown in Figure \ref{frame}.
\begin{align}
h^{m}_{\mathit{BOW}} = \mathit{CAM}(h^{m}_{\mathit{BOW}},h_{\mathit{PLM}}^{m}),m\in\{r,a\}
\end{align}

\subsubsection{Visual Information Embedding.}

For visual information embedding, we follow \cite{lin2022multi} and adopt the pre-trained visual model (PVM) to learn visual representation. For consistency, we use the embeddings generated by \cite{lin2022multi}. The visual representation is sent through a feed-forward layer to get final visual embedding $h^v$.
\begin{equation}
h^{v} = W_v \cdot \mathit{PVM}(v_i)+b_v
\end{equation}
where $W_v$ and $b_v$ are learnable parameters.

\subsubsection{Joint Embedding}

Inspired by recent works \cite{lin2022multi, chen2022multi, chen2023meaformer}, we implement a simple weighted concatenation by integrating the multi-modal features into a joint representation $h^j$:
\begin{equation}
\setlength\abovedisplayskip{3pt}
\setlength\belowdisplayskip{3pt}
h^j = \mathop{Concat}\limits_m\big[ softmax(\alpha^m)\cdot h^m\big]
\label{joint}
\end{equation}
where $m$ denotes the modal type and $m\in\{g, r_{\mathit{BOW}}, r_{\mathit{PLM}}, a_{\mathit{BOW}}, a_{\mathit{PLM}}, v\}$,  which means that $h^m $$\in$$ \{h^g, h^r_{\mathit{BOW}},h^r_{\mathit{PLM}},h^a_{\mathit{BOW}},h^a_{\mathit{PLM}}, h^v\}$. $\alpha^m$ is the trainable attention weight for the modality of $m$. $Concat$ means concatenation operation.

\begin{table*}[h!]
	\renewcommand\arraystretch{1}
	\fontsize{9}{10}
	\centering
	\begin{tabular}{@{}c|c|cccc|cccc@{}}
		\toprule
		\multirow{2}{*}{\textbf{Seeds}}              & \multirow{2}{*}{\textbf{Model}} & \multicolumn{4}{c|}{\textbf{FB15K-DB15K}}                             & \multicolumn{4}{c}{\textbf{FB15K-YAGO15K}}                            \\
		&                                 & $Hits@1$          & $Hits@5$          & $Hits@10$         & $\mathit{MRR}$             & $Hits@1$          & $Hits@5$         & $Hits@10$         & $\mathit{MRR}$             \\ \midrule
		\multirow{12}{*}{20\%} & PoE                             & 0.1260          & -               & 0.2510          & 0.1700          & 0.2500          & -               & 0.4950          & 0.3340          \\
		& MMEA                            & 0.2648          & 0.4513          & 0.5411          & 0.3570          & 0.2339          & 0.3976          & 0.4800          & 0.3170          \\
		& EVA                             & 0.1340          & -                & 0.3380          & 0.2010          & 0.0980          & -               & 0.2760          & 0.1580          \\
		& ACK-MMEA                        & 0.3040          & -               & 0.5490          & 0.3870          & 0.2890          & -               & 0.4960          & 0.3600          \\
		& MSNEA                           & \underline{0.6527}    & \underline{0.7685}    & \underline{0.8121}    & \underline{0.7080}    & {0.4429}    & {0.6255}    & {0.6983}    & {0.5290}    \\
		& MCLEA                           & 0.4450          & -               & 0.7050          & 0.5340          & 0.3880          & -               & 0.6410          & 0.4740          \\
		& MultiJAF                        & 0.4800          & 0.5760          & 0.6010          & 0.5230          & \underline{0.4630}          & \underline{0.6580}          & \underline{0.7310}          & \underline{0.5540}          \\
		& MEAformer                       & 0.4340          & -               & 0.7280          & 0.5340          & 0.3250          & -               & 0.5980          & 0.4160          \\
		& MEAformer(+)                    & 0.5780          & -               & 0.8120          & 0.6610          & 0.4440          & -               & 0.6920          & 0.5290          \\
		& PCMEA(ours)                            & \textbf{0.6763} & \textbf{0.8214} & \textbf{0.8872} & \textbf{0.7280} & \textbf{0.5896} & \textbf{0.7518} & \textbf{0.8347} & \textbf{0.6460} \\ \midrule
		\multirow{9}{*}{50\%}  & PoE                             & 0.4640          & -               & 0.6580          & 0.5330          & 0.4110          & -               & 0.6690          & 0.4980          \\
		& MMEA                            & 0.4165          & 0.6210          & 0.7035          & 0.5120          & 0.4026          & 0.5723          & 0.6451          & 0.4860          \\
		& EVA                             & 0.2230          & -               & 0.4710          & 0.3070          & 0.2400          & -               & 0.4770          & 0.3210          \\
		& ACK-MMEA                        & 0.5600          & -               & 0.7360          & 0.6240          & 0.5350          & -               & 0.6990          & 0.5930          \\
		& MCLEA                           & 0.5730          & -               & 0.8000          & 0.6520          & 0.5430          & -               & 0.7590          & 0.6160          \\
		& MEAformer                       & 0.6250          & -               & 0.8470          & 0.7040          & 0.5600          & -               & 0.7800         & 0.6400          \\
		& MEAformer(+)                    & \underline{0.6900}    & {-}         & \underline{0.8710}    & \underline{0.7550}    & \underline{0.6120}    & {-}         & \underline{0.8080}    & \underline{0.6820}    \\
		& PCMEA(ours)                              & \textbf{0.7375} & \textbf{0.8614} & \textbf{0.9154} & \textbf{0.7810} & \textbf{0.6702} & \textbf{0.8151} & \textbf{0.8857} & \textbf{0.7210} \\ \midrule
		\multirow{9}{*}{80\%}  & PoE                             & 0.6660          & -               & 0.8200          & 0.7210          & 0.4920          & -               & 0.7050          & 0.5720          \\
		& MMEA                            & 0.5903          & 0.8041          & 0.8687          & 0.6850          & 0.5976          & 0.7849          & 0.8389          & 0.6820          \\
		& EVA                             & 0.3700          & -               & 0.5850          & 0.4440          & 0.3940          & -               & 0.6130          & 0.4710          \\
		& ACK-MMEA                        & 0.6820          & -               & 0.8740          & 0.7520          & 0.6760          & -               & 0.8640          & 0.7440          \\
		& MCLEA                           & 0.7300          & -               & 0.8830          & 0.7840          & 0.6530          & -               & 0.8350          & 0.7150          \\
		& MEAformer                       & 0.7730          & -               & 0.9180          & 0.8250          & 0.7050          & -               & 0.8740           & 0.7680          \\
		& MEAformer(+)                    & \underline{0.7840}    & -           & \underline{0.9210}    & \underline{0.8340}    & \underline{0.7240}    & {-}         & \underline{0.8800}    & \underline{0.7830}    \\
		& PCMEA(ours)                              & \textbf{0.8204} & \textbf{0.9232} & \textbf{0.9644} & \textbf{0.8580} & \textbf{0.7556} & \textbf{0.8819} & \textbf{0.9424} & \textbf{0.8020} \\ \bottomrule
	\end{tabular}
	
	\caption{Main experiments on FB15K-DB15K and FB15K-YAGO15K  with different proportions of entity alignment seeds. The best results are highlighted in bold and the second best results are underlined. The ``-" denotes that the results are not available, and the ``+" means the iterative results.}
	\label{mainresult}
	
\end{table*}

\subsection{Mutual Information Enhanced Cross-modality Alignment}

Different modality information enriches the entity representation from different perspectives. The joint embedding generated from fusion mechanism is more comprehensive than uni-modal embedding. Thus, aligning uni-modal embedding with the joint embedding can transfer the knowledge from the joint embedding back to uni-modal ones, resulting in better uni-modal representation. Concretely, we minimize the Align Loss ($\mathcal {L}_{AL} $) to reduce the difference between uni-modality and joint modality and to realize the knowledge transfer. For aligned pair $(e_1^i, e_2^i)$:
\begin{align}
\small
\mathcal {L} _{AL}^m =  -\sum_m &E_{i \in B} \left[D_\mathit{KL}(\mathbb{Q}_{h^j}(e_1^i, e_2^i) || \mathbb{Q}_{h^m}(e_1^i, e_2^i))\right. \nonumber \\[-2mm]
&\left. + D_\mathit{KL}(\mathbb{Q}_{h^j}(e_2^i,e_1^i)||\mathbb{Q}_{h^m}(e_2^i,e_1^i))\right]
\end{align}
where $D_\mathit{KL}(\cdot)$ represents the Kullback-Leibler Divergence, $\mathbb{Q}_{h^j}(e_1^i, e_2^i) = h^j_{e_1^i} \otimes h^j_{e_2^i}$ and $h^j_{e_1^i}$ is the joint embedding of $e_1^i$. $\mathbb{Q}_{h^m}(e_1^i, e_2^i)$ is calculated in an analogous way using the uni-modal embedding $h^m$, $ m\in \{g, r_{\mathit{BOW}},  r_{\mathit{PLM}}, a_{\mathit{BOW}}, a_{\mathit{PLM}}, v\}$.

However, direct alignment of cross-modality only encourages integrating information from different modalities while mixing the noise from each modality irrelevant to our task. Thus, we use the mutual information estimator MINE \cite{belghazi2018mutual} to enhance mutual information, which can be utilized to mine the modal-invariant information between different modalities and filter out modality-specific random noise \cite{qi2023ssmi, bao2023wukong}. Specifically, we maximize the MI between joint embedding $h^j$ and uni-modal embedding $h^m$:
\begin{flalign}
I(\widehat{h^j, h^m}) = &\mathit{max}\; I( h^j;  h^m) \nonumber \\
= & \mathit{max}\; D_{\mathit{KL}}(\mathbb{P}_{h^jh^m} \;  || \; \mathbb{P}_{h^j} \otimes \mathbb{P}_{h^m}  ) \nonumber\\
= & \mathit{sup} \; \mathbb{E}_{\mathbb{P}_{h^jh^m}} [\Phi]- log(\ \mathbb{E}_{\mathbb{P}_{h^j} \otimes \mathbb{P}_{h^m}})[e^\Phi]
\end{flalign}
where $\mathbb{P}_{h^jh^m}$ represents joint distribution, $\mathbb{P}_{h^j}$ and $\mathbb{P}_{h^m}$ are marginal distributions of joint embedding $h^j$ and uni-modal embedding $h^m$, respectively. $\mathit{sup} $ represents supremum function and $\Phi$ is a simple nonlinear layer. Specifically, in the multi-modal entity alignment task, the loss for MI maximization is:
\setlength\abovedisplayskip{1pt}
\setlength\belowdisplayskip{1pt}
\begin{equation}
\mathcal {L}_{MI}= - \sum_m I(\widehat{h^j, h^m})
\end{equation}
where $h^m \in \{h^g,h^r_{PLM}, h^a_{PLM}, h^v\}$ is uni-modal entity representation, $h^j$ denotes the joint-modal embedding. 

\subsection{Contrastive Learning with Pseudo-Label Calibration}

In this section, contrastive learning with pseudo-label calibration strategy is designed for semi-supervised EA. It mainly consists of two parts: (1) \textit{Pseudo-label calibration} provides more reliable pseudo-aligned entity pairs to expand the training data set and decrease error propagation. (2) \textit{Momentum-based contrastive learning mechanism} can be more effective in pulling the aligned pairs closer and pushing unaligned pairs away.
\subsubsection{Pseudo-label Calibration.}
Ensemble learning always plays a crucial role in improving prediction performance and overcoming the model homogenization issues of single model. Accordingly, we devise a pseudo-label calibration strategy that improves the confidence of pseudo-aligned pairs via the modal ensemble. Specially, we introduce a dynamic prediction dictionary, where the prediction of the present epoch will be stored. After $\omega$ epochs, the new prediction will first be compared with the previously stored results. For simplicity, we set $\omega$ to 2. If the two results are identical, the sample, as well as the predicted sample, will be classified as pseudo-label. Otherwise, the new prediction will be used to update the dictionary.

In addition, we introduce a data reordering method to accelerate model learning and optimize the quality of pseudo-label generation. In the early stage of model training, we rearrange the order of the labeled data, which is used to speed up the convergence of the model and capture crucial features. We first calculate the cosine similarity based on the entity joint representation $h^j$ and then put together data items with higher cosine similarity. Therefore, we put the more similar data in one mini-batch that the model can fit the general features and make the model initially distinguishable. 

\subsubsection{Momentum-Based Contrastive Learning.}

Recent studies \cite{zeng2022interactive, liu2022selfkg} have shown the popularity of contrastive learning in entity alignment. The aligned seeds can be naturally regarded as positive samples, whereas any non-aligned pairs can be considered as negative samples due to the convention of 1-to-1 alignment constraint. Formally, for the $i$-th entity $e_1^i \in E_1$ of mini-batch $B$, the positive set is defined as $P^i = \{e_2^i | e_2^i \in E_2\}$, where $(e_1^i, e_2^i)$ is an aligned pair. The negative set includes two parts, inner-graph unaligned pairs from the source KG $G_1$ and cross-graph unaligned pairs from the target KG $G_2$, defined as $N_1^i = \{e_1^i | \forall e_1^j \in E_1, i \ne j \}$ and $N_2^i = \{e_2^j | \forall e_2^j \in E_2, i \ne j \}$, respectively. Since contrastive learning and pseudo-label calibration are performed simultaneously, the scale of the alignment seeds is gradually expanded as training progressively proceeds, and the corresponding positive and negative sets are dynamically updated.

Motivated by MoCo \cite{he2020momentum} and Fast-MoCo \cite{ci2022fast}, momentum-based contrastive learning methods adopt an asymmetric forward path, and the two encoded samples from two paths (online path and target path) form a pair for contrastive learning, which has been proven to be effective in many scenarios. In this work, we propose to apply momentum-based contrastive learning to the field of semi-supervised entity alignment. On the online path, online entity representation is generated by the online encoder. On the target path, momentum entity representation is generated by a slowly moving momentum encoder. Thus, for the $i$-th entity $e_1^i \in E_1$ of mini-batch $B$, its representation is generated by the online encoder, and the representations of its positive set $P^i$ and negative set $N^i $ are generated by momentum encoder. While the online encoder’s parameter $\theta_{online}$ is instantly updated with the back-propagation, the target encoder’s parameter $\theta_{target}$ is asynchronously updated with momentum by:
\begin{equation}
\theta_{target} \leftarrow 	\kappa \cdot \theta_{target} + (1-	\kappa) \cdot \theta_{online}, 	\kappa\in [0,1)
\end{equation}

To be specific, we define the alignment probability distribution $q_m(e_1^i, e_2^i)$ of the modality $m$ for each positive pair $(e_1^i, e_2^i)$ as:
\begin{align}
\small
&q_m(e_1^i, e_2^i) = \nonumber\\[-2.5mm]
&\frac{\delta_m(e_1^i, e_2^i)}{\delta_m(e_1^i, e_2^i) +\! \sum\limits_{e_1^j \in N_1^i}\delta_m(e_1^i, e_1^j)+ \!\sum\limits_{e_2^j \in N_2^i}\delta_m(e_1^i, e_2^j)}
\setlength\abovedisplayskip{3pt}
\end{align}
where $\delta_m(u,v) = exp(f_m(u)^Tg_m(v)/\tau)$, $f_m(\cdot)$ and $g_m(\cdot)$ are the online encoder and the momentum encoder of the modality $m$, respectively. $T$ denotes transpose operation and $\tau$ is a temperature parameter. Notably, the distribution is directional and asymmetric for each input; the distribution for another direction is thus defined similarly as $q_m(e_2^i, e_1^i)$. The loss function of contrastive learning can be calculated by: 
\setlength\abovedisplayskip{0pt}
\setlength\belowdisplayskip{0pt}
\begin{equation}
\mathcal{L}^m_{CL} = -E_{i \in B}log\Big[\frac{1}{2}(q_m(e_1^i, e_2^i)+q_m(e_2^i, e_1^i))\Big]
\end{equation}

We employ stage-wise momentum-based contrastive learning loss on uni-modal and joint representation. Specifically, in the first stage, only the online network is trained and updated. In the second stage, the momentum network is initialized and trained simultaneously with the online network. The time of changing the training strategy is $ts$, and the time span for updating the momentum network is $\rho$ epochs.

\subsection{Optimization Objective}
The overall loss of the PCMEA is given below,
\setlength\abovedisplayskip{0pt}
\setlength\belowdisplayskip{0pt}
\begin{equation}
\mathcal{L} = \sum_{m\in M1} \mathcal{L}_{AL}^m + \mathcal{L}_{MI} + \sum_{m\in M2}\mathcal{L}^m_{CL}
\end{equation}
where $M1=\{g, r_{\mathit{BOW}}, r_{\mathit{PLM}},a_{\mathit{BOW}},a_{\mathit{PLM}}, v\}, M2=M1\cup\{h^j\}$, and $h^j$ denotes the joint embedding in Eq. (\ref{joint}).

\begin{figure}[htbp]
	\centering
	\includegraphics[width=0.78\linewidth]{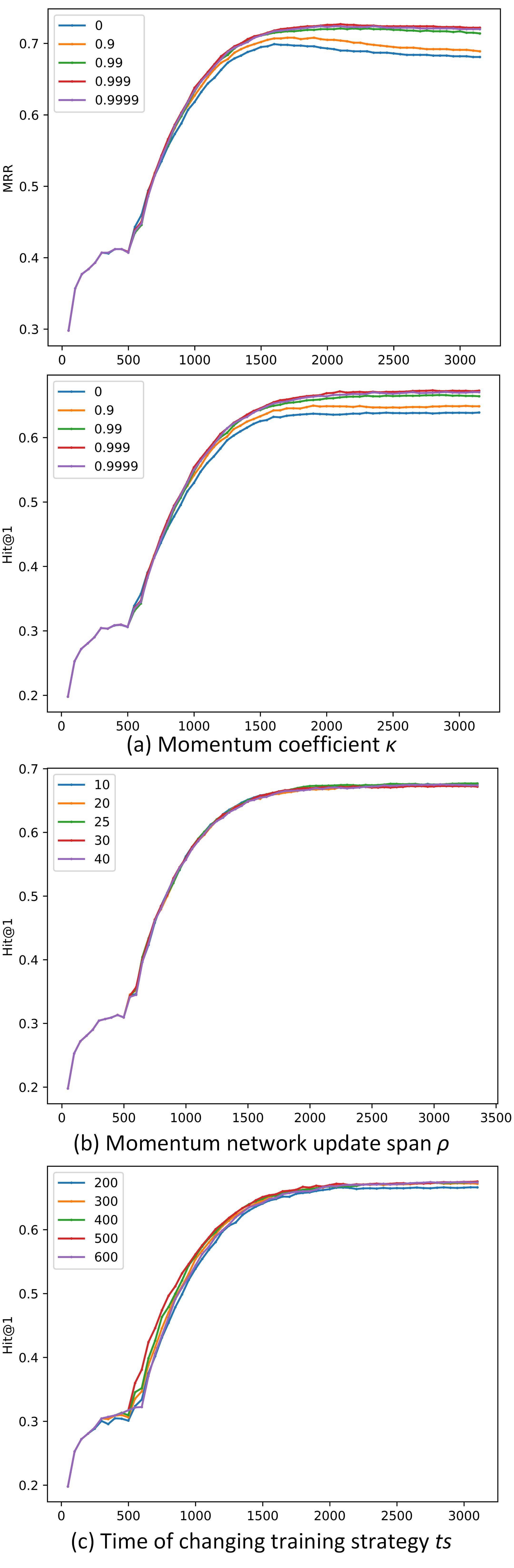}
	\caption{Study on (a) momentum coefficient, (b) momentum network update span, (c) time of changing training strategy.}
	\label{train_strategy}
\end{figure}

\section{Experiments}
\subsection{Experimental Settings}
Two cross-KG EA datasets are adopted for evaluation, including FB15K-DB15K and FB15K-YAGO15K, which are the most representative datasets in MMEA task \cite{chen2020mmea,chen2022multi,lin2022multi}. FB15K is one of the most widely used data sets in the field of link prediction. Entities from DBpedia and YAGO aligned with FB15K are extracted through the SameAs links, which are utilized to build DB15K and YAGO15K datasets.

Our baseline model is MCLEA \cite{lin2022multi} and we also compare our method against other 7 state-of-the-art multi-modal EA methods, which can be classified into three categories: 1) traditional multi-modal EA methods, including PoE \cite{liu2019mmkg}, MMEA \cite{chen2020mmea}, and EVA \cite{liu2021visual}; 2) multi-modal EA method based on pre-trained language model, such as ACK-MMEA \cite{li2023attribute}; 3) multi-modal EA method based on contrastive learning, including MSNEA \cite{chen2022multi}, MultiJAF  \cite{cheng2022multijaf} and MEAFormer \cite{chen2023meaformer}. For all baselines, we report the original results from their literature.

Our model is implemented based on Pytorch, an open-source deep learning framework. The pre-trained language models (Bert \cite{kenton2019bert}, T5 \cite{raffel2020exploring}, RoBerta \cite{liu2019roberta}, Albert \cite{lan2019albert} ChatGLM-6B \cite{du2022glm} and LLaMA-7B \cite{touvron2023llama}) are downloaded from Hugging Face\footnote{https://huggingface.co/} and all of them are base version. All experiments were conducted on a server with two GPUs (NVIDIA-SMI 3090).

\subsection{Results}

To verify the effectiveness of our method, we report overall average results in Table \ref{mainresult}. It shows performance comparisons on FB15K-DB15K and FB15K-YAGO15K datasets with different splits on training/testing data of alignment seeds, i.e., 2:8, 5:5, and 8:2.

From Table \ref{mainresult}, we can observe that: 1) Our model outperforms all the baseline of MMEA methods in terms of all metrics on both datasets. Particularly, our model brings about 9.04\%-23.13\% (16.21\% on average) improvement on FB15K-DB15K and 10.26\%-20.16\% (14.38\% on average) on FB15K-YAGO15K in terms of $Hits@1$ for all proportions of training data over baseline MCLEA. The superiority of our method demonstrates that the proposed structure and training strategy substantially boosts performance in the semi-supervised settings. 2) Our model shows a clear improvement over the traditional multi-modal EA method, which reveals that a more rational multi-modal information fusion method, as well as an appropriate training strategy, can make full use of the available data. 3) Our model is significantly more effective than other EA methods based on PLM, suggesting that PLM is useful for embedding relations and attributes but needs to be coupled with the neighborhood structure of the entity. 4) Moreover, our method clearly surpasses the state-of-the-art baseline by 2.36\% in $\mathit{Hits}@1$ and 2\% in $\mathit{MRR}$ on FB15K-DB15K and significantly outperforms the best baseline by 12.66\% in $\mathit{Hits}@1$ and 9.2\% in $\mathit{MRR}$ on FB15K-YAGO15K based on 20\% aligned seeds. The above results also indicate that our modified contrastive learning strategy and model structure are superior.

\begin{table}[h!]
	\fontsize{9}{8}
	\centering
	\begin{tabular}{@{}lccc@{}}
		\toprule
		\multirow{2}{*}{\textbf{Variants}} & \multicolumn{3}{c}{\textbf{FB15K-DB15K}}                            \\
		& \textbf{$Hits@1$}  & \textbf{$Hits@10$} & \textbf{$\mathit{MRR}$} \\ \midrule
		\textbf{Ours}                      & \textbf{0.6763}                    & \textbf{0.8872}           & \textbf{0.7280}      \\ \midrule
		w/o Image                          &0.6510                     &0.8740            &0.7070        \\
		w/o Bag-of-words                   & 0.5303                    & 0.7582           & 0.5860       \\
		w/o PLM                            & 0.6502                 & 0.8746           & 0.7050       \\
		w/o Cross-attention                & 0.6521                    & 0.8719           & 0.7070       \\
		w/o MI Loss                        & 0.6718                   & 0.8841           & 0.7230       \\
		w/o Align Loss                     & 0.6674                   & 0.8818           & 0.7200       \\
		w/o CL Loss                        & 0.3610                   & 0.7204           & 0.4320       \\
		w/o LC1                            & 0.6647       &0.8809   & 0.7190 \\
		w/o LC2                            & 0.6727        &0.8753      &0.7220      \\
		w/o LC1\&LC2                       &0.6588          &0.8706      &0.7120    \\ \midrule
		\multicolumn{4}{c}{Attribute Embedding  using different PLM}                                             \\ \midrule
		Bert                               & 0.6682                   & 0.8803           & 0.7190       \\
		\textbf{Roberta}                   & \textbf{0.6763}                    & \textbf{0.8872}           & \textbf{0.7280 }      \\
		Albert                             & 0.6634                  & \underline{0.8811}           & 0.7180       \\
		T5                                 & 0.6718                   & \underline{0.8811}           & \underline{0.7230}       \\
		ChatGLM-6B                             &0.6650           &0.8788      &0.7170    \\
		LLaMA-7B                               &\underline{0.6746}    &0.8787  &\underline{0.7230}   \\
		\midrule
		\multicolumn{4}{c}{Relation Embedding  using different PLM}                                              \\ \midrule
		Bert                               & 0.6706                 & 0.8853           & 0.7230       \\
		Roberta                            & 0.6693                  & \textbf{0.8952}           & 0.7260       \\
		Albert                             & 0.6672                & 0.8828          & 0.7200        \\
		\textbf{T5}                                 & \underline{0.6763}        & \underline{0.8872}         &\textbf{ 0.7280 }      \\
		ChatGLM-6B                              &0.6726             &0.8825       &0.7230   \\
		LLaMA-7B                               &\textbf{0.6768}     &0.8829      &\underline{0.7270}   \\
		\bottomrule
	\end{tabular}
	\caption{Variant experiments on FB15K-DB15K (20\%). w/o means removing the corresponding module from the complete model. LC1 and LC2 mean dynamic prediction dictionary and the reordering method in pseudo-label calibration strategy, respectively. $\mathit{In}$ represents the maximum sequence length input to LLMs.}
	\label{ablation}
	
\end{table}

\subsection{Ablation Study}

To investigate the effectiveness of each module in PCMEA, we perform variant experiments, whose results are shown in Table \ref{ablation}. From Table \ref{ablation}, we notice that: 1) The impact of the bag-of-words method (BOW) tends to be more significant than PLM on encoding relations and attributes. When combining PLM and BOW, the cross-attention mechanism must be used to bring out the power of PLM. We believe this is because the semantic differences between the vectors produced by the two encoders are obvious, and the attention mechanism can reduce the semantic gap. 2) The removal of the contrastive learning loss (CL loss) has the greatest impact with respect to the removal of the other two loss functions, since CL loss directly clusters together similar entities, while the others transfer information between different modalities. 3) Removing pseudo-label calibration drops 0.36\%-1.75\% in $\mathit{Hits}@1$, showing that improving pseudo-labeling quality is necessary for semi-supervised contrastive learning and can contribute to model performance. 4) We analyze the influence of embedding models by replacing different pre-trained language models. Specially, we test four PLMs (Bert et al.) and two large language models (ChatGLM-6B and LLaMA-7B). The results show that different embedding models affect entity alignment to a certain extent, and stronger model can improve performance. 5) The variants without image modality decline on all metrics, which hints that the multi-modal information and rational utilization are necessary for the EA task.

\textbf{Impact of hyper-parameters.} We conduct hyper-parameter studies using FB15K-DB15K, showing results in Figure \ref{train_strategy}. The main hyper-parameters in our method are momentum coefficient $\kappa$, momentum network update span $\rho$, and time of changing training strategy $ts$. For momentum coefficient $\kappa$, a proper large $\kappa$ (e.g. 0.999) bring better  stability and accuracy in $Hits@1$ and $\mathit{MRR}$, illustrating momentum-based contrast learning is more effective than just contrast learning. Varying time span $\rho$ shows little difference. For time $ts$ of changing training strategy, time $ts$ obvious effects the surge in $\mathit{Hits}@1$ after changing the strategy, with $ts = 500$ allowing faster convergence. Besides, $ts$ barely affects post-convergence performance.

\section{Conclusion}
In this work, we propose a semi-supervised pseudo-label calibration multi-modal entity alignment framework named PCMEA. It utilizes various embedding methods and attention mechanisms to obtain multi-modal entity representation. Instead of direct interaction and fusion of multi-modal embedding, we apply mutual information maximization to filter out task-independent noise and transfer cross-modality information. To boost the quality of pseudo-label and contrastive learning, we combine pseudo-label calibration with momentum-based contrastive learning, which helps pull aligned pairs closer and improve alignment performance. Experimental results show that PCMEA can consistently outperform prior state-of-the-art methods, producing high-quality alignment performance even under 20\% labeled data settings.

\section{Acknowledgments}
This work is partially supported by the National Natural Science Foundation of China under Grant No. 61772534, and partially supported by Public Computing Cloud, Renmin University of China.

\bibliography{aaai24}

\end{document}